\documentclass[twocolumn]{IEEEtran}
\usepackage{lscape}
\usepackage{booktabs}
\usepackage{authblk}
\usepackage{url}
\usepackage{silence}
\usepackage{subfigure}
\usepackage[font=scriptsize]{caption}

\usepackage[pdftex]{graphicx}
\usepackage{epstopdf}
\usepackage{pifont}
\usepackage[table,xcdraw]{xcolor}
\usepackage{array}
\usepackage{verse}
\usepackage{longtable}
\usepackage{supertabular}
\usepackage{cite}
\usepackage{color}
\usepackage{amsmath}
\usepackage{lettrine}
\usepackage{hyperref}
\usepackage{multirow}
\usepackage{multicol}
\usepackage{dirtytalk}
\usepackage{color}
\usepackage{bm}
\usepackage{algorithm}
\usepackage{algorithmic}
\DeclareMathOperator*{\argmax}{arg\,max}
\DeclareMathOperator*{\argmin}{arg\,min}

\begin{document}
\title{An Active Learning Method for Diabetic Retinopathy Classification with Uncertainty Quantification}
\author{Muhammad Ahtazaz Ahsan$^1$, Adnan Qayyum$^1$, Junaid Qadir$^1$, and Adeel Razi$^{2,3}$ \\ \vspace{2mm}
$^1$ Information Technology University (ITU), Punjab, Lahore, Pakistan\\
$^2$ Turner Institute for Brain and Mental Health, Monash University, Clayton, Victoria, Australia \\ 
$^3$ Wellcome Centre for Human Neuroimaging, UCL, London, United Kingdom}
\maketitle

\begin{abstract}
In recent years, deep learning (DL) techniques have provided state-of-the-art performance on different medical imaging tasks. However, the availability of good quality annotated medical data is very challenging due to involved time constraints and the availability of expert annotators, e.g., radiologists. In addition, DL is data-hungry and their training requires extensive computational resources. Another problem with DL is their black-box nature and lack of transparency on its inner working which inhibits causal understanding and reasoning. In this paper, we jointly address these challenges by proposing a hybrid model, which uses a Bayesian convolutional neural network (BCNN) for uncertainty quantification, and an active learning approach for annotating the unlabelled data. The BCNN is used as a feature descriptor and these features are then used for training a model, in an active learning setting. We evaluate the proposed framework for diabetic retinopathy classification problem and have achieved state-of-the-art performance in terms of different metrics.
\end{abstract}
\section{Introduction}
\label{sec:introduction}
Recent advancements in machine learning (ML) techniques, in particular, deep learning (DL) based methods have achieved state-of-the-art performance in many complex medical imaging tasks such as image classification \cite{afshar2018brain}, segmentation \cite{fraz2012ensemble}, annotation \cite{yan2018deeplesion}, and retrieval \cite{qayyum2017medical}. However, to learn a better representation of the underlying distribution of data, DL requires large-scale training data. However, the availability of large amount of clinical data is a real challenge due to various ethical, monetary and privacy constraints. In addition, the annotation of medical data is a very costly, and time-consuming, task. This motivates the development of DL approaches that can learn from limited medical data or that can incorporate both annotated and unannotated data.

Another issue plaguing DL is that even when trained on large-scale (training) datasets, DL is a black-box method that lack underlying mechanistic understanding and has inherent issues that make it uncertain about the predictions made. In DL-empowered healthcare, a few key challenges are noticeable that make quantification of uncertainty difficult. Bengoli et al. \cite{begoli2019need} described three such challenges that include; (i) lack of well-understood laws for clinical data, unlike the physical world which is supported by well-defined mathematical laws; (ii) absence of causal co-relation between the inputs and outputs of the DL model (the absence of a causal relationship limits the conclusion that is drawn from a DL model); (iii) imperfections that are embedded in the data which makes a DL model uncertain of its prediction. Moreover, real-world data also contain missing elements that demand specialized methods for data imputation and uncertainty quantification. To overcome these challenges, a model should be carefully developed by considering the efficiency challenges and uncertainty, especially for clinical applications. In this regard, CNNs with Bayesian inference are more useful and reliable rather than using deterministic CNNs which lack the quantification of uncertainty.

Diabetic retinopathy (DR) is a neuropathic complication arising from damage to the retinal optic nerve that can lead to blindness. DR deteriorates (due to neurodegenrative and microvasculopathic factors \cite{muc2018associations}) over time if left untreated and therefore, early detection is of utmost importance to avoid irreversible damage to vision. There are many diseases that are associated with DR such as retinal vascular closure, abnormal vessel growth, and diabetic macular edema. Each disease has its own unique pathophysiology that is crucial for the diagnosis and prognosis, the resulting complexity means increased risk of inaccurate diagnosis and treatment (possibly due to human error or fatigue). These complications can lead to undesired circumstances and can cause visual damage. On the other hand, an automatic DR detection and classification system can assist the clinicians in their routine clinical work by predicting and locating the possible disease which at the same time decreases the risk of human error that may arise due to misinterpretation, fatigue, and tiredness.

In this paper, we provide a unified framework for simultaneously addressing the problems of uncertainty quantification, training with limited labeled data and leveraging unlabelled data. The following are the specific contributions of this paper. 
 \begin{enumerate}
     \item We propose a hybrid model that consists of two key components: (i) a Bayesian CNN descriptor module to address the uncertainty problem and (ii) an active learning (AL) module to train the model with unlabelled data. 
     \item We integrate and evaluate two AL approaches (pool-based sampling and query by committee) for training the model in an AL environment.
     \item We extensively evaluate the proposed hybrid model for the DR classification task with uncertainty quantification using different performance metrics and as well as for the task of uncertainty quantification.
 \end{enumerate}
 
\textit{Organization of the paper:} A brief background of related terminologies is presented in Section \ref{sec:back}. An overview of related work focused on DR classification is presented in Section \ref{sec:related work}. A detailed explanation of our proposed methodology is presented in Section \ref{sec:methodology}. The dataset description and implementation details are discussed in Section \ref{sec:data and experiments}. A detailed analysis of results and some future research issues are provided in Section \ref{sec:results and discussion}. Finally, the paper is concluded in Section \ref{sec:conclusion}.
\section{Methods}
\label{sec:back}
\subsection{Bayesian Inference}
Deducing model parameters or properties about a probability distribution from data is referred to as inference. Bayesian inference uses Bayes' theorem to update the probability distribution upon the availability of new data. The classical Bayes rule comprises three components: (i) prior distribution (also known as beliefs), (ii) posterior distribution, and (iii) likelihood. The prior distribution is typically assumed as a normal distribution (with some mean and standard deviation) or as a Gaussian process.
 
Deep neural network (DNN) is a linear combination of weights and bias vectors followed by a non linear operation, e.g., \textit{ReLU}, \textit{tanh}, or \textit{sigmoid} as activation function applied on linear output vector. At each epoch, the loss function e.g., cross-entropy loss in case of multi-class classification is optimized by backpropagating the loss through the neural network using an optimizer, (e.g., SGD or Adam). Applying the Bayes rule on the weights and biases of a neural network allows us to update them over a distribution rather than a single real number (as done in conventional DL model training). The Bayesian inference estimates the posterior distribution by examining all the possible outcomes of each new training instance. An example is described below for further explanation.  

Let's assume we have a labeled dataset, $D=\left \{ x_{n},y_{n}\right \}$, where $x_n$ denotes samples and $y_n$ are their corresponding labels. The Bayes rule for estimating posterior distribution over the network latent parameters $w$ can be mathematically defined as:

\begin{equation}\label{eq:erl}
\mathit{p(w|D)=\dfrac{p(D|w)\times p(w)}{p(D)}},
\end{equation}
where, $p(w)$ is prior, $p(D|w)$ is the likelihood and $p(w|D)$ is the posterior. The posterior distribution $p(w|D)$ is approximated by  minimizing the Kullback-Liebler (KL) divergence between the prior and variational distribution $q(w|\theta)$ \cite{sun2019functional,blundell2015weight}.

\begin{equation}\label{eq:el2}
    \theta^{*}= \argmin_{\theta} KL[q(w|\theta)||p(w)] - E_{q_{w|\theta}}\left [ \log p(D|w) \right]
\end{equation}

Equation \ref{eq:el2} is a cost function which is known as variational free energy, which is an expected lower bound on the (log) model evidence, and it is solved as an optimization problem when we parameterize the weights $w$ over a parameter $\theta$ for $q(w|\theta)$. By assuming (conjugate) Gaussian prior and posterior (known as Laplace approximation) which is fully factorized such as to approximate the posterior by minimizing the KL-divergence loss is known as mean-field variational inference.

\subsection{The Problem of Uncertainty in DL}
Uncertainty in a DL model can be defined as how much a model is unsure about its prediction \cite{shridhar2019comprehensive}. Uncertainty quantifies the entropy or surprise of the model on unseen data and it can be classified into two categories, (i) \textit{aleatoric uncertainty} and (ii) \textit{epistemic uncertainty} \cite{der2009aleatory}. \textit{Aleatoric uncertainty} is due to the noisy or unclean data and it is inherently present in the data. On the other hand, \textit{epistemic uncertainty}, also known as model uncertainty is the amount of uncertainty in the DL model. Aleatoric uncertainty is modeled by assuming a prior over the set of weights given the training data and epistemic uncertainty is modeled by placing a distribution over the output of the model. 

\subsection{Active Learning}
The AL framework is built on the hypothesis articulated in \cite{settles2009active} that if the learning algorithm is allowed to choose the data from which it learns, it will perform better with less training. AL-based model (also known as the learner) query the label of only that sample on which it find it difficult to classify. The difficulty is quantified using a query function \textit{aka} query strategy. Learner selects that sample from a large-scaled unlabelled dataset, queries its label, and the unlabelled data is augmented into already known data for training.

There are three different scenarios which are used to query the unlabelled instance, which are: (i) membership query synthesis; (ii) stream-based selective sampling; and (iii) pool-based sampling. In membership query synthesis, the model tries to construct the new data samples based on some underlying distribution. In stream-based selective sampling, it is assumed that acquiring a label for each data instance of unlabelled data is free. The learner must decide which instance it needs to query. We have used the pool-based sampling and query-by-committee in our proposed model, which are described below.
 
\subsubsection{Pool-Based Sampling}
The most commonly used and intuitive framework of pool-based query strategy is uncertainty sampling \cite{lewis1994sequential}. Uncertainty sampling uses different mathematical functions to measure the uncertainty, named as, least confident sampling, margin sampling, and entropy sampling. In the first method, the learner only queries that instance for which it is least confident for assigning it a label. For example, in a multi-class classification problem, for any sample data, let's assume $x$, and the associated label $y$, and $\theta$ represents the weights of trained model, the uncertainty sampling can be mathematically defined as:

\begin{equation}
\label{eqn:LQ}
    x_{LC}^{*} = \argmax_{\theta}(1-P(y^{'}|x)),
\end{equation}

where, for a multi-class classification problem $y^{'} = \argmax_{y} 1-p_{\theta}(y|x)$, as explained in \cite{settles2009active}. The least confident sampling can be understood by an example. Suppose, you have two instances to classify and each instance can have three possible labels. So the class probabilities for first instance are $[0.3,0.4,0.3]$ and for second instance are $[0.4,0.45,0.15]$. Selecting the most likely labels for these two instances would give the values of $0.4$ and $0.45$ and subtracting these probability values from 1 and then taking the maximum value from the result will query for the first instance.

The second method for uncertainty sampling is the margin sampling which incorporates the posterior of the second most likely label, and thus it solves the shortcoming of the least confident sampling, which only gives the single most likely label. The margin sampling can be mathematically defined as:

 \begin{equation}
     x_{M}^{*}=\argmin_{x} p_{\theta}(y_{1}^{'}|x)-p_{\theta}(y_{2}^{'}|x)
 \end{equation}
Again taking same example of classification as in least confident sampling, the margin sampling does the following thing. The difference between the first and the second most likely label for the first instance is $0.1$ and for the second instance is $0.05$. Therefore, the learner will select the second instance as it has the smaller margin value.

 The last method is the entropy sampling \cite{settles2009active} which uses the entropy as query strategy function. Entropy sampling can be defined mathematically as:
 \begin{equation}
 \label{eqn:ent}
      \mathit{x_{H}^{*} = \argmax_{x}-\sum_{i=1}^{c}p_{\theta}(y_{i}|x)\log(p_{\theta}(y_{i}|x)),}
 \end{equation}
where $c$ represents the total number of classes. Quoting the similar example, the entropy values for the first instance is $1.57$ and for the second instance is $1.46$. So the learner will choose the first instance on which it has the maximum value of entropy.
\subsubsection{Query-by-Committee}
In a query-by-committee (QBC) setting, a ``committee'' of two or more classifiers is formed. Each committee member is assigned a subset of currently available training data. After being trained on training data, each member maintains its own hypothesis. Each committee member votes on the label of a candidate example after being trained on the data available to it and that instance is selected for querying the label on which the committee members disagree. The main objective of QBC approach is to minimize the version space, which is defined as, the set of hypotheses that are consistent with the currently available data. That also means that all hypotheses of different models agree on the labeled data points but they disagree on some unlabelled data points and these points lie in the uncertain region. In this way, the QBC approach query the candidate sample in the most uncertain region. 
 
The measurement of disagreement in committee based sampling can be computed using vote entropy sampling \cite{settles2009active} which can be mathematically defined as:

 \begin{equation}
 \label{eqn:ves}
     x_{VE}^{*} = \argmax_{x}-\sum_{i} \frac{V(y_{i})}{C}\log\frac{V(y_{i})}{C},
 \end{equation}
where, $V(y_{i})$ is the number of votes obtained by the label $y_{i}$ and $C$ denotes the size of committee.
\section{Related Work}
\label{sec:related work}
DR is one of the co-morbidity associated with diabetic patients that can cause blindness. In the literature, significant research has focused on DR classification. The state of the art in DR classification mainly relies on DL-based decision support systems. In this section, we present the overview of the related literature. We start by first discussing the development of grading systems in DR, then we discuss the DR classification problem using DL, then we discuss risk assessment of other diseases associated with DR, and finally, we discuss the use of AL-based methods for leveraging the limited annotated data for supervised classification task.

\subsection{Multi-level DR Research}
DR is generally classified into four or five different grading levels based on the disease severity. Wilkinson et al. \cite{wilkinson2003proposed} proposed five classes of DR, while Yun et al. \cite{yun2008identification} proposed four classes of DR. In five class schema, DR is divided into five classes based on the severity of the disease which are (i) no DR, (ii) mild, (iii) moderate, (iv) severe, and (v) proliferative.

In the four class schema, the normal or no DR class is merged with mild DR class with the other classes being moderate, severe, and proliferative, respectively.

\subsection{DR Classification}
In the literature, different approaches for DR classification has been presented that mainly rely on ML-based techniques. 
For instance, Roychowdhury et al. \cite{roychowdhury2013dream} used the classical ML algorithms like KNN, SVM, and GMMs to perform binary classification, i.e., DR or No DR. In \cite{priya2013diagnosis}, authors have analyzed the DR classification problem using probabilistic neural networks (PNNs), support vector machine (SVM), and Bayes classifier. 
In addition to traditional ML-based approaches, DL-based methods have also been proposed for DR classification. For instance, Gulshan et al. proposed a DL-based DR classification algorithm that uses the Inception-V3 model pre-trained on ImageNet \cite{gulshan2016development}. Yang et al. \cite{yang2017lesion} proposed a two-stage DR classification algorithm that performs two tasks, DR classification and lesions localization in the retinal fundus images. A method named machine learning bagging ensemble classifier (ML-BEC) is proposed in \cite{somasundaram2017machine}, which extracts different features (e.g., features related blood vessels, optic nerve, neural tissue, disk size, and thickness, etc.) for DR classification using-stochastic ML model. The use of a simple neural network, backpropagation neural network, and convolutional neural network for the DR classification is presented in \cite{dutta2018classification}. The use of VGG19 for the DR classification is presented in \cite{mateen2019fundus}, the authors also used the combination of Gaussian mixture models, dimensionality reduction techniques like principal component analysis (PCA), and singular value decomposition techniques (SVD) for performing the DR classification task. In a similar study, Gadekallu et al. proposed the use of deterministic CNNs and transitional ML models for DR classification along with different data pre-processing and dimensionality reduction techniques \cite{gadekallu2020early}. In \cite{khalifa2019deep}, authors evaluated different DL models like AlexNet, VGGNet, GoogleNet, SqueezeNet, and ResNet with transfer learning for  multi-class DR classification. Chetoui et al. \cite{chetoui2020explainable} proposed to use EfficientNet-B7 DL model for DR classification. 
\subsection{Bayesian DL in DR Classification}
In the literature, various studies have investigated the use of Bayesian models for estimating model uncertainty that is trained for the task of DR classification. For instance, Hani et al. investigated the use of the Gaussian Bayes classifier and v-fold cross-validation (VFCF) for DR classification \cite{hani2010gaussian}. Sedai et al. \cite{sedai2018joint} proposed a method that exploits different layers of retinal images pixel by pixel for quantifying uncertainty in DR images by using the Bayesian DL approaches. Filos et al. \cite{filos2019systematic} proposed a systematic comparison of Bayesian DL benchmarks like ensemble model, ensemble dropout, Monte-Carlo dropout, and mean-field variational inference for the DR classification task. They re-formulated the multi-class classification problem to a binary classification problem. Ranganath et al. \cite{krishnan2020specifying} proposed a method for selecting informed weight prior comprising two stages, i.e., firstly, they find the maximum likelihood estimate of weights by using DNN and then setting up the weight prior for using empirical Bayes.
\subsection{Active Learning for Medical Data Analysis}
Although AL is not a new technique but still it works considerably better than semi-supervised learning in most real-world problems for handling unlabelled data. Wang et al. \cite{wang2016cost} proposed a cost-effective method for image classification by training a DL model in the AL setting. 
Gal et al. \cite{gal2017deep} proposed a technique that uses the combination of Bayesian DL and AL by designing a special acquisition function, also known as the query strategy. They evaluated the proposed method for skin cancer diagnosis. Haut et al. \cite{haut2018active} used the hyperspectral image classification using a DL model in an AL setting and trained the model on limited labeled samples to achieve a good classification performance for labeling large unannotated data.
\section{Methodology}
\label{sec:methodology}
\subsection{The Proposed Model}
\begin{figure}[!t]
\centering
\includegraphics[width=1\linewidth]{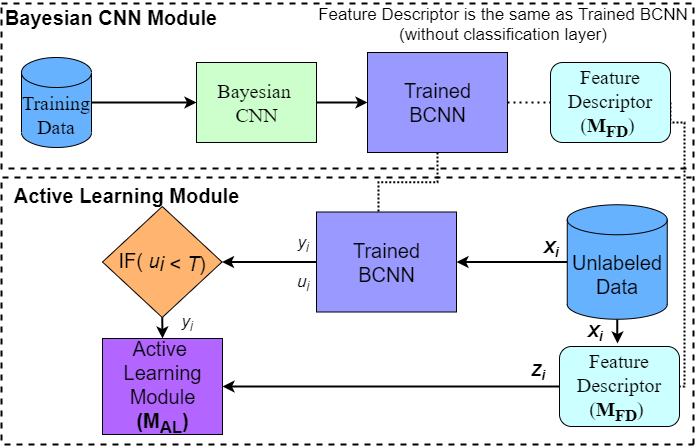}
\caption{An illustration of the proposed hybrid model that has two components, i.e., Bayesian CNN module as a feature descriptor and active learning module. Image $\boldsymbol{X_{i}}$ is the queried image, $\boldsymbol{Z_{i}}$ is the Bayesian feature vector, T is the threshold value of uncertainty, and $y_{i}$ is the predicted label.}
\label{fig:workflow}
\end{figure}

Our proposed hybrid model has two main components, i.e., Bayesian CNN module and active learning module, as shown in Figure \ref{fig:workflow}. We use Bayesian CNN module as feature descriptor by extracting the output of a parametric layer. The active learning module picks an image $\boldsymbol{X_{i}}$ from the unlabelled data and puts a request to the trained Bayesian CNN module for its label $y_{i}$, uncertainty $u_{i}$, and $\boldsymbol{Z_{i}}$ as its respective feature vector. If the uncertainty is less than the threshold value of $T$ (details are in Section \ref{subsec:UQ}), the label is forwarded to the active learning module (details are described in Section \ref{sec:data and experiments}).
\subsection{Problem Formulation}
\label{subsec:prob_form}
We assumed that we have both labeled and unlabelled samples of retinal images for training our supervised learning model. Therefore, we divide the whole dataset into three disjoint parts (with no overlapping), i.e., training, validation, and testing sets. Suppose the dataset $D_{T_r} =\left \{ \boldsymbol{X_{i}},y_{i}\right \}$ where $i= 1,2,3,...,N$ represents the training dataset, ${D_V =\left \{ \boldsymbol{X_{j}},y_{j}\right \}}$ where $j=1,2,3,...,M$ represents the validation dataset, and ${D_T =\left \{\boldsymbol{X_{k}},y_{k}\right \}}$ where $k=1,2,3,...,S$ represents the test dataset respectively. We further divide the $D_{V}$ into two categories, named as limited labelled dataset $V_{L}$ and large pool of unlabelled dataset $V_{U}$, respectively, which are used for training the active learning model. The Bayesian CNN model is represented by $M_{FD}$ and AL model is represented by $M_{AL}$, respectively.
After training $M_{FD}$, we extracted feature vectors $\boldsymbol{Z_{k}}$ as test data for AL, $\boldsymbol{Z_{V_L}}$ as limited labeled data, and $\boldsymbol{Z_{V_U}}$ as a large pool of unlabelled data from trained feature descriptor $M_{FD}$. These features are then used to train and evaluate $M_{AL}$. An unlabelled sample $V_{U}$ is queried by $M_{AL}$. Our main objective is to optimize $M_{AL}$ in such a way that it leverages the unlabelled data by getting a label from $M_{FD}$ on which it is confident so that $M_{AL}$ achieves an increase in performance after adding unlabelled data points into the training.

In our proposed hybrid model, we integrate the two AL approaches of pool-based sampling and query-by-committee sampling (described in the previous section). The algorithm for pool-based sampling and query-by-committee sampling can be seen in Algorithm \ref{algo2} and Algorithm \ref{algo3}, respectively.
\section{Data and Experiments}
\label{sec:data and experiments}
\subsection{Data Description}
\begin{figure}[!t]
\centering
\includegraphics[width=0.95\linewidth]{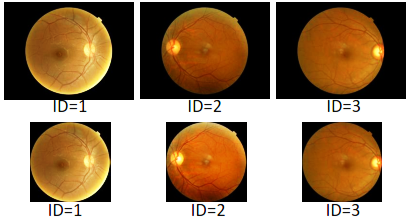}
\caption{Images of different dimensions in original data set (from ID=1 to ID=3) are centre-cropped to produce the dimensions of 224 $\times$ 224. }
\label{img:fov}
\end{figure}

We used \textit{APTOS2019} \cite{aptos2019} dataset which contains 3662 high-resolution color retinal images annotated into five classes (i.e., $1805$, $370$, $999$, $193$, and $295$ samples for No DR, Mild DR, Moderate DR, Severe DR, and Proliferative DR, respectively). These high-resolution images contain surrounding black patches around the corner of the images. We have carefully analyzed the dimensions of these images and cropped them according to their field of view (FOV), an example of this cropping is shown in Figure \ref{img:fov}. After cropping the images, we have resized all images to the square size of $224\times224\times3$.


The original dataset was not large enough to train a generalized DL model. We enlarged the size of the dataset up to $4 \times$ the original dataset . We used different data augmentation techniques for binary and multi-class classification to augment the training sets (the details are in the later sections).



\begin{algorithm}[H]
\caption{Pool-Based Sampling}
\textbf{Input:}  Limited Labeled Data $\boldsymbol{Z_{V_{L}}}$, unlabelled Pool Data Features $\boldsymbol{Z_{V_{U}}}$, Test Data Features $\boldsymbol{Z_{D_{T}}}$, Number of Queries $Q$, Query instances $i$, and Number of Epochs $E$\\
\textbf{Output:} Predicted Label on Test Data Features $\boldsymbol{Z_{D_{T}}}$
\begin{algorithmic}[1]
\FOR{$e \in E$}
    \STATE \textbf{Train} $M_{PBS}$ Using $\boldsymbol{Z_{V_{L}}}$
\ENDFOR
\STATE $q$=0
\WHILE{$\boldsymbol{Z_{V_{U}}}$ is not \textit{empty} \OR q $\leq$ $Q$}
    \STATE \textit{Select} $i$ $\in$ $\boldsymbol{Z_{V_{U}}}$ 
    \STATE Query $i$ samples using $M_{PBS}$
    \STATE Assign class label to  i samples using $M_{FD}$
    \STATE $\boldsymbol{Z_{V_{L}}} \leftarrow  \boldsymbol{Z_{V_{L}}} \bigcup \boldsymbol{Z_{V_{U_i}}}$ 
    \STATE \textit{Retrain} $M_{PBS}$ Using $\boldsymbol{Z_{V_{L}}}$ \\
    \STATE Increment $q$ by 1
\ENDWHILE

\FOR{ $\boldsymbol{X} \in \boldsymbol{Z_{D_{T}}}$}
    \STATE $y$ = $\argmax$ $M_{PBS}(\boldsymbol{X})$ \\
\ENDFOR
\end{algorithmic}
\label{algo2}
\end{algorithm}

\begin{algorithm}[H]
\caption{Query-By-Committee Sampling}
\textbf{Input:}  Limited Labeled Data $\boldsymbol{Z_{V_{L}}}$, unlabelled Pool Data Features $\boldsymbol{Z_{V_{U}}}$, Test Data Features $\boldsymbol{Z_{D_{T}}}$, Number of Queries $Q$, Number of Epochs $E$, Committee Members $M$\\
\textbf{Output:} Predicted Label on Test Data $\boldsymbol{Z_{D_{T}}}$
\begin{algorithmic}[1]
\FOR{$m \in M$}
 \FOR {$e \in E$}
    \STATE \textbf{Select} $j$ disjoint samples $\forall$ $\boldsymbol{Z_{j}}$
    \STATE \textbf{Train} $M_{QBC_m}$ Using $\boldsymbol{Z_{j}}$
 \ENDFOR
\ENDFOR
\STATE q=0
\WHILE{$\boldsymbol{Z_{V_{U}}}$ is not \textit{empty} \OR $q \leq Q$}
    \STATE \textit{Select} $k \in \boldsymbol{Z_{V_{U}}}$
    \STATE Perform Consensus 
    \STATE Assign class label to $k$ samples using $M_{FD}$
    \STATE $\boldsymbol{{Z_{V_{k}}}} \leftarrow  \boldsymbol{Z_{V_{k}}} \bigcup \boldsymbol{Z_{V_{U_k}}}$
    \STATE \textit{Retrain} $M_{QBC_m}$ Using $\boldsymbol{Z_{V_{k}}}$
\ENDWHILE
\FOR{ $\boldsymbol{X} \in \boldsymbol{Z_{D_{T}}}$}
 \FOR{$m \in M$}
    \STATE $y_m = M_{QBC_m}(\boldsymbol{X})$ \\
 \ENDFOR
  \STATE $y = \argmax \frac{1}{M}\sum_{m=1}^{M}y_m$
\ENDFOR
\end{algorithmic}
\label{algo3}
\end{algorithm}

\subsubsection{Data Description for Binary Classification}
For binary classification experimental evaluation, we have simplified the multi-class classification problem to the binary class classification problem, i.e., class 0/1 classification where class 0 represents instances of \textit{No DR (NDR)} and class 1 represents those having \textit{DR}, a similar formulation was also followed in \cite{roychowdhury2013dream,filos2019systematic,lam2018automated}. Keeping this class merging in mind and to avoid biased training, we assign the class weight of 1 to NDR and class weight of 4 to DR. The class merged dataset was then divided into $D_{T_{r}}$, $D_{V}$, and $D_{T}$ with the sizes of 9076, 1181, and 1212, respectively.

\subsubsection{Data Description for Multi-Class Classification}
For multi-class classification, the data augmentation techniques like \textit{vertical flip} and \textit{random brightness} up to a range of 20\% are applied. To avoid the class imbalance problem, classes with less number of samples are augmented more, i.e., oversampling less samples class. To incorporate the class imbalance further, we compute class weights for the $i^{th}$ class using Eq. \ref{eq:cw}.

\begin{equation}
\mathit{x_{i}=\frac{\left|D_{T_r}\right|}{C \times \left|y_{i}\right|}},
\label{eq:cw}
\end{equation}
where $D{_{T_{r}}}$ represents training data, $y_{i}$ are total instances of a any class $i$, and $C$ denotes the total number of classes in the dataset. The dataset for multi-class classification has also been divided into $D_{T_{r}}$, $D_{V}$, and $D_{T}$ having $8940$, $1915$, and $1920$, respectively. 

\begin{figure*}[!t]
\centering
\includegraphics[width=1\linewidth]{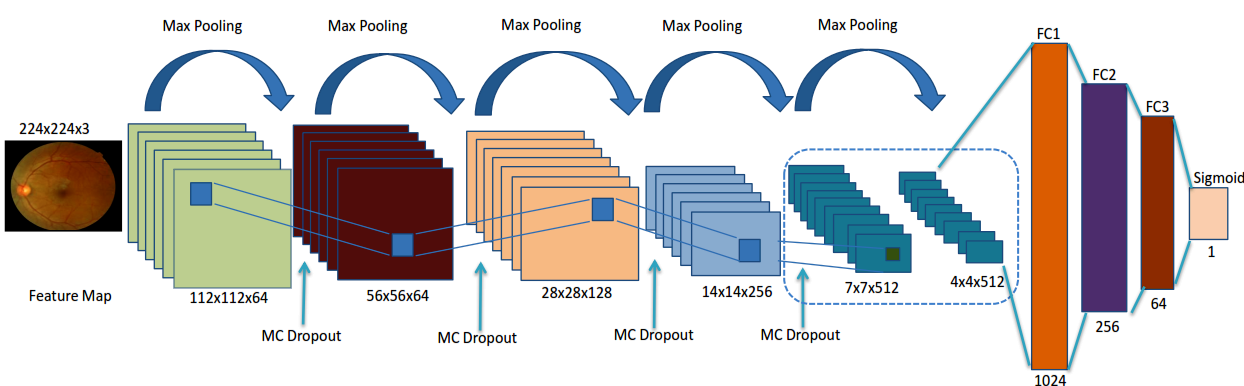}
\caption{Diagram of CNN. Dropout layer is added after each convolutional layer, whereas feature map shows the output of each block after applying max-pooling where dropout is applied after each convolutional layer in the block.Also, FC1, FC2, FC3, and Sigmoid layer has 1024, 256, 64, and 1 neurons respectively.}
\label{fig:cnn_bin}
\end{figure*}

\subsection{Model Architecture}
\subsubsection{Model Architecture for Binary Classification}
We used a VGG-like CNN architecture, as proposed by \cite{filos2019systematic} with some modifications. We used the \textit{Monte-Carlo (MC)} drop out (a method for realizing Bayesian inference) after each parametric layer in BCNN and a simple drop out after each block in CNN. We use an initial number of base filters to be 64 and increased the filter size as shown in Figure \ref{fig:cnn_bin}.

\subsubsection{Model Architecture for Multi-Class Classification}
The model architecture for multi-class classification is the same as shown in Figure \ref{fig:cnn_bin} except some modification that are described as follows. The initial filter size is set to be 32 and these filters are increased in the multiple of 2 in every block. In multi-class model, the \textit{Batch Normalization}, and the \textit{Dropout} layers are used simultaneously in every block. Also, \textit{LeakyReLU} (which helps in minimizing the diminishing gradients effect) is used instead of simple \textit{ReLU}. The number of neurons in  FC1, FC2, and FC3 are set to be 2048, 512, and 128, respectively. Finally, the sigmoid layer is replaced with the \textit{Softmax} layer having $5$ neurons in it (i.e., five number of classes). We have performed all of our experiments using \textit{TensorFlow 2.0.} along with \textit{Keras} as its API.

\subsection{Training Feature Descriptor Module}
\subsubsection{Details for Binary Classification}
We trained both models (i.e., CNN and BCNN) on the dataset for $15$ epochs, at learning rate of $0.0001$, batch size of $128$, and the MC dropout rate of $0.10$ using \textit{Adam} optimizer. For BCNN model, the dropout is applied in both training and testing time (to realize the Bayesian inference), along with the $L_{2}$ regularization of $0.0001$ in each weighted layer to reduce the over-fitting.

\subsubsection{Details for Multi-Class Classification}
In multi-class classifications, the batch size is set to be $64$ and the number of epochs are set to be $80$. Learning rate, $L_{2}$ regularization value and the optimizer are the same as used in binary classification. For training our models, we used the dropout rate of $0.5$ for CNN and $0.20$ for BCNN, respectively.

\subsection{Implementation Details for Active Learning Module}
Once the feature descriptor model i.e. $M_{FD}$ is trained, we extracted the features, i.e., $Z_{k}$, $\boldsymbol{Z_{V_{L}}}$ and, $\boldsymbol{Z_{V_{U}}}$ from the last convolution layer of each model. The reason for selecting the last convolutional layer is to get the large-sized features which can be used for training the $M_{PBS}$ or $M_{QBC}$ in active learning settings. For pool-based sampling, we initially trained $M_{PBS}$ with $100$ samples. For  query-by-committee, we initially selected three committee members, and each committee member is initially trained on $100$ disjoint samples.

\subsubsection{Training AL Model for Binary Classification}
We extracted the features from the last convolutional layer which had the output of $7\times7\times512$ taken out from the $M_{FD}$. For pool-based sampling, $10$ number of samples from $\boldsymbol{Z_{V_{U}}}$ are returned by the $M_{PBS}$. $M_{FD}$ returns the most likely label along with the uncertainty (in measure of entropy). These samples are augmented with $\boldsymbol{Z_{V_{L}}}$ and the $M_{PBS}$ is retrained. We have queried a total of 50 times from the $M_{PBS}$ and 500 newly labeled samples into the $\boldsymbol{Z_{V_{L}}}$. Similarly, for the query-by-committee approach, these $10$ newly labeled samples are added to each known training dataset $\boldsymbol{Z_{V_{L}}}$ for all three $M_{QBC}$, and the models are retrained. Also, in the query-by-committee approach, the only difference in the experiments is that after training all three committee members, the prediction (for testing purpose) is done by taking the average of class probabilities and picking the most likely label.

\begin{figure*}[!h]
    \centering
    \captionsetup{justification=centering}
    \subfigure[]{\includegraphics[width=0.4\textwidth]{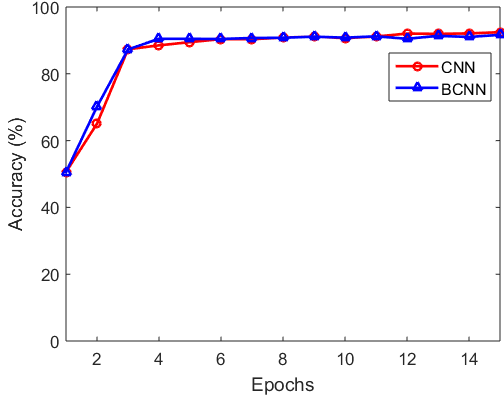}\label{fig:acc}}
    \subfigure[]{\includegraphics[width=0.4\textwidth]{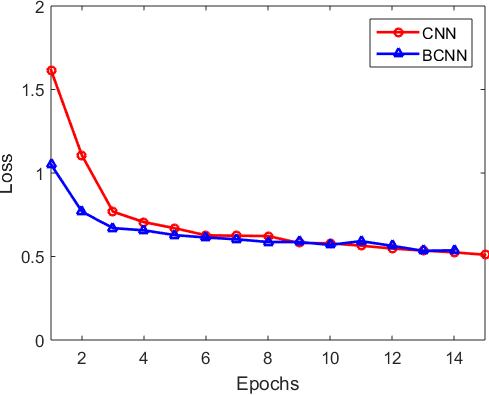}\label{fig:loss}}
    \caption{The depiction of models (i.e., CNN and BCNN) performance in terms of (a) accuracy and (b) loss.}
    \label{fig:cnn_bcnn}
\end{figure*}

\subsubsection{Training AL Model for Multi-class Classification}
For multi-class classification, we also extracted the feature vectors having the dimension of $10\times10\times256$ from the last convolutional layer of $M_{FD}$. For $M_{PBS}$ and $M_{QBC}$, the initial number of training data samples are kept to $100$ and $300$, respectively. The $300$ samples for the query-by-committee approach are distributed to three committee members with a size of $100$ samples. $16$ samples from $\boldsymbol{Z_{V_{U}}}$ are augmented to $\boldsymbol{Z_{V_{L}}}$ after querying these models. A total of $1200$ samples are queried in training both $M_{PBS}$ and $M_{QBC}$ models. We use standard categorical cross-entropy loss and focal loss \cite{lin2017focal}. Focal loss was introduced by the \textit{Facebook AI research group} and was initially proposed for dense object detection purpose. Focal loss is mathematically defined for the cross-entropy loss in Eq. \ref{eqn:floss}, where $\alpha$ is the weighting factor and $\gamma$ is the modulating factor. We selected these values to be 4 and 2 $\alpha$ = 4 worked best for our case  (as suggested by \cite{lin2017focal}). 

\begin{equation}
\label{eqn:floss}
    \mathit{F_{L}= -\alpha (1-p_t)^{\gamma}\log(p_t)}
\end{equation}

We trained our models in such a way that standard categorical cross-entropy loss is applied in training $M_{PBS}$ and $M_{QBC}$ and stored the weights of these models respectively. Then we change the loss function and used the pre-trained weights. We did this to focus more on the examples that are being added in the training data $\boldsymbol{Z_{V_{L}}}$ after querying from $\boldsymbol{Z_{V_{U}}}$.

\subsection{Uncertainty Quantification (UQ)}
\label{subsec:UQ}
The \textit{Mont-Carlo (MC) dropout} method is a way of implmenting the Bayesian inference from CNNs while dropout is also enabled at inference time. MC samples are the number of feed-forward passes for a single image. After applying MC iterations, the most likely label is obtained by averaging the class probabilities. This method (known as MC-dropout) is applied in our proposed approach.

\subsubsection{UQ for Binary Classification}
We assume that the predictive entropy value greater than or equal to $0.5$ represents high uncertainty and low confidence while entropy less than $0.5$ is assumed to represent low uncertainty and high confidence value (a similar assumption is made in \cite{filos2019systematic}).
\subsubsection{UQ for Multi-class Classification}
We start by assuming the case when our model $M_{FD}$ is giving equal probabilities to all classes which gave the entropy value of $2.32$. Like in binary class classification, we set the threshold value of $1.276$ which is 55\% of the maximum threshold value. Varying amount of Monte-Carlo samples are selected and the results are reported in Section \ref{sec:results and discussion}.

\section{Results and Discussions}
\label{sec:results and discussion}
In this section, we present our results for both binary and multi-class classification. Moreover, a detailed comparison with state of the art methods is also presented in this section.
\subsection{Results for Binary Classification}
\label{sub:res_bin}
\subsubsection{Training Feature Descriptor}
In CNN, \textit{aka} simple or deterministic CNN, the dropout layers are disabled at the time of inference, whereas in the Bayesian CNN, the dropout is enabled at the inference time. The learning curves for training CNN and BCNN are shown in Figure \ref{fig:cnn_bcnn}. Figure \ref{fig:acc} is representing the training accuracy and Figure \ref{fig:loss} is depicting the training loss over the number of epochs. It can be observed from the figures that accuracy is increasing while the loss is decreasing over the increase in number of epochs.  The learning curves are sort of similar in behavior as the same model parameters are being trained on the same data. As explained earlier, the key difference is in the inference time. In training the feature descriptor, the key idea is to take out a feature vector that can be a faithful representative of the true posterior distribution, which is achieved in Bayesian CNN instead of simple CNN.

The classification performance report for simple CNN and Bayesian CNN on validation data for binary classification task is given in the Table \ref{tab:one}.

\begin{table}[!h]
\centering
\caption{Comparison of CNN and Bayesian CNN Models on validation data for Binary Classification.}
\begin{tabular}{|c|c|c|c|c|}
\hline
\multicolumn{5}{|c|}{\textbf{Class 0/ No DR}} \\ \hline
Model &
  \begin{tabular}[c]{@{}c@{}}Accuracy\\ (\%)\end{tabular} &
  \begin{tabular}[c]{@{}c@{}}Precision\\ (\%)\end{tabular} &
  \begin{tabular}[c]{@{}c@{}}Recall\\ (\%)\end{tabular} &
  \begin{tabular}[c]{@{}c@{}}F1-Score\\ (\%)\end{tabular} \\ \hline
CNN &
  93 &
  96 &
  90 &
  93 \\ \hline
BCNN &
  94 &
  95 &
  95 &
  94 \\ \hline
\multicolumn{5}{|c|}{\textbf{Class 1/DR}} \\ \hline
Model &
  \begin{tabular}[c]{@{}c@{}}Accuracy\\ (\%)\end{tabular} &
  \begin{tabular}[c]{@{}c@{}}Precision\\ (\%)\end{tabular} &
  \begin{tabular}[c]{@{}c@{}}Recall\\ (\%)\end{tabular} &
  \begin{tabular}[c]{@{}c@{}}F1-Score\\ (\%)\end{tabular} \\ \hline
CNN &
  93 &
  90 &
  96 &
  93 \\ \hline
BCNN &
  94 &
  94 &
  95 &
  94 \\ \hline
\end{tabular}
\label{tab:one}
\end{table}

\subsubsection{Comparison of Active Learning Methods}
\begin{figure*}[!h]
\centering
\subfigure[Pool-based sampling]{\includegraphics[width=0.4\textwidth]{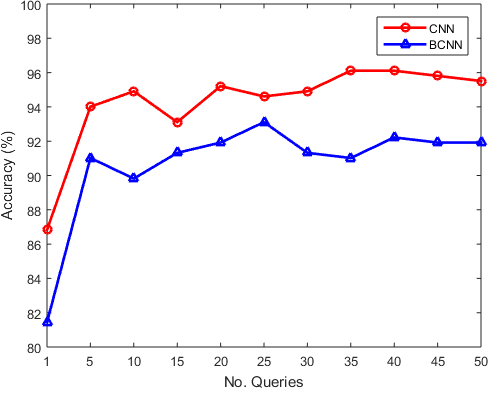}\label{fig:pbs}}
\subfigure[Query by Committee sampling ]{\includegraphics[width=0.4\textwidth]{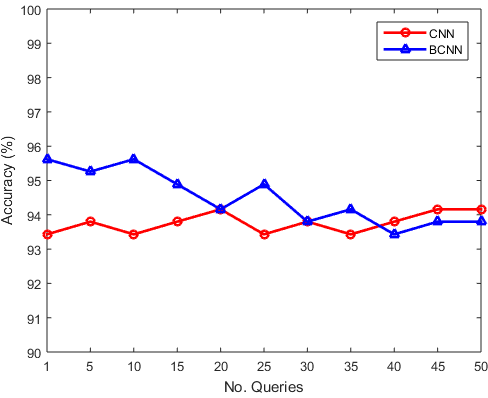}\label{fig:qbc}}
\caption{Comparison of active learning approaches in binary classification for CNN and BCNN using (a) pool-based sampling and (b) query by committee sampling.}
\label{fig:al_bin}
\end{figure*}

In pool-based sampling model, i.e., $M_{PBS}$, both the CNN and BCNN model's accuracy is increased to a certain level over the number of queries and when the model is trained enough, the accuracy over the test data has become stable. The final accuracy that the model achieved is 94\% and 91\% after the $50$ queries have been reached for CNN and BCNN, respectively as depicted in Figure \ref{fig:pbs}.
The performance of the query-by-committee model is shown in Figure \ref{fig:qbc}. The initial accuracy of the three committee members is 54\%, 56\%, and 59\%, respectively. The final accuracy for both CNN and BCNN in query-by-committee is around 93\%. While comparing the pool-based sampling and query-by-committee, the CNN model is achieving higher performance than the BCNN. One reason for this behavior is the enabling of dropout in both $M_{FD}$ and $M_{PBS}$. Suggesting that there is a trade-off between performance \textit{Vs.} accurate uncertainty quantification.


\begin{figure*}[!h]
    \centering
    \captionsetup{justification=centering}
    \subfigure[]{\includegraphics[width=0.4\textwidth]{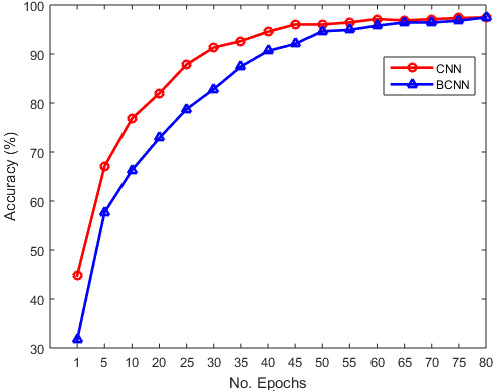}\label{fig:acc_mc}}
    \subfigure[]{\includegraphics[width=0.4\textwidth]{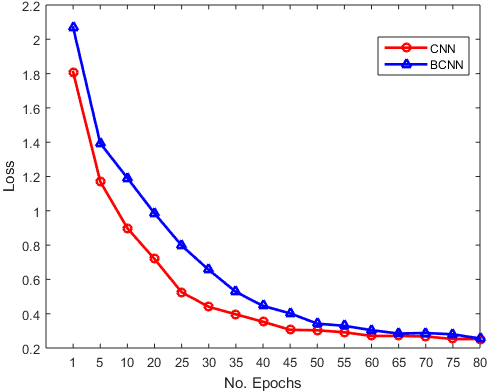}\label{fig:mc}}
    \caption{The depiction of models (i.e., CNN and BCNN) performance for multi-class classification in terms of (a) accuracy and (b) loss.}
    \label{fig:mc_cnn_bcnn}
\end{figure*}

\subsubsection{UQ for Binary Classification}
For the uncertainty quantification (UQ), we perform a series of extensive experiments. We perform these experiments by changing the number of MC samples. Our experiments show that our model is predicting the class of \textit{No DR} and \textit{DR} with more confidence and the wrongly classified or the correctly classified samples with more uncertainty are less in numbers. To reduce the cost of time and computations, we only reported our results of the model's uncertainty up to 50 MC samples.
Table \ref{tab:mc_perf} shows three different MC samples, and the model is tested with a dropout rate of 0.10, while the dropout is enabled at inference time. 

\begin{table}[!h]
\centering
\caption{Analysis of Bayesian CNN results on binary classification.}
\begin{tabular}{|p{45mm}|c|c|c|}
\hline
\multicolumn{1}{|c|}{\textbf{Analysis}} & \textbf{MC-5} & \textbf{MC-20} & \textbf{MC-50} \\ \hline
Correctly classified ($u$ \textless \textit{T})       & 979 & 978 & 976 \\ \hline
Correctly classified ($u$ \textgreater{}= \textit{T}) & 161 & 165 & 163 \\ \hline
Wrongly classified ($u$ \textless \textit{T})        & 13  & 13  & 12  \\ \hline
Wrongly classified ($u$ \textgreater{}= \textit{T})  & 59  & 56  & 61  \\ \hline
\end{tabular}
\label{tab:mc_perf}
\end{table}


Table \ref{tab:mc_com} shows that UQ has identified those rare cases which have been misclassified with uncertainty greater than the threshold value of uncertainty, which is 0.5 in our case, and those instances which are correctly classified by the model with the uncertainty greater than the threshold value. This represents a trade-off between performance versus uncertainty quantification. In case, if we only consider the true positives with the uncertainty less than the threshold value, our proposed framework achieves 81\% of the accuracy (94\%) which is 13\% less than the deterministic CNN.      




\subsection{Results for Multi Class Classification}
\label{sub:res_mul}
\subsubsection{Training Feature Descriptor}
The learning curves of the feature descriptor module for multi-class classification are shown in Figure \ref{fig:mc_cnn_bcnn}. Both CNN and BCNN are achieving up to $95\%$ accuracy on training data when trained for $80$ epochs. Similar kind of insights can be drawn by observing these curves (as we observed for binary classification). The performance of CNN is slightly higher than the BCNN due to less information blocking in CNN architecture. Classification report for multi-class classification in terms of accuracy, precision, recall and F1-score performed on test data is given in Table \ref{tab:mc_cr}. The confusion matrix for BCNN and CNN are also given below in Figure \ref{fig:confm_mc}.

\begin{figure}[!h]
\centering
\captionsetup{justification=centering}
\includegraphics[width=0.95\linewidth]{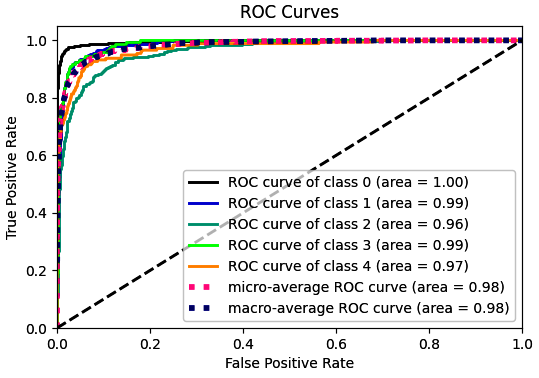}
\caption{ROC curves for all classes which are showing area under the curve (AUC) approximately equal to one.}
\label{fig:roc_mc}
\end{figure}

We also reported the ROC curves for the test data for the BCNN model by applying the dropout rate of $0.30$ and the number of MC iterations to $25$ in Figure \ref{fig:roc_mc}. These ROC curves and class-wise area under the curve (AUC) show that the BCNN model is providing good estimation of posterior distribution, which lacks in simple CNN models.

\begin{table*}[!h]
\caption{Comparison of Different Monte-Carlo Samples with the uncertainty quantification results.}
\centering
\begin{tabular}{|c|c|c|c|c|c|c|c|}
\hline
\multicolumn{2}{|c|}{\textbf{Dropout=0.10}} &
  \multicolumn{2}{c|}{\textbf{MC-5}} &
  \multicolumn{2}{c|}{\textbf{MC-20}} &
  \multicolumn{2}{c|}{\textbf{MC-50}} \\ \hline
\textbf{Original   Label} &
  \textbf{Predicted Label} &
  \textbf{Entropy \textless 0.5} &
  \textbf{Entropy \textgreater{}= 0.5} &
  \textbf{Entropy \textless 0.5} &
  \textbf{Entropy \textgreater{}= 0.5} &
  \textbf{Entropy \textless 0.5} &
  \textbf{Entropy \textgreater{}= 0.5} \\ \hline
No DR & No DR    & 422 & 138 & 422 & 142 & 420 & 140 \\ \hline
No DR & DR    & 10  & 29  & 10  & 25  & 10  & 29  \\ \hline
DR    & No DR & 3   & 30  & 3   & 31  & 2   & 32  \\ \hline
DR    & DR    & 557 & 23  & 556 & 23  & 556 & 23  \\ \hline
\end{tabular}
\label{tab:mc_com}
\end{table*}

\begin{figure*}[]
    \centering
    \captionsetup{justification=centering}
    \subfigure[CNN]{\includegraphics[width=0.45\textwidth]{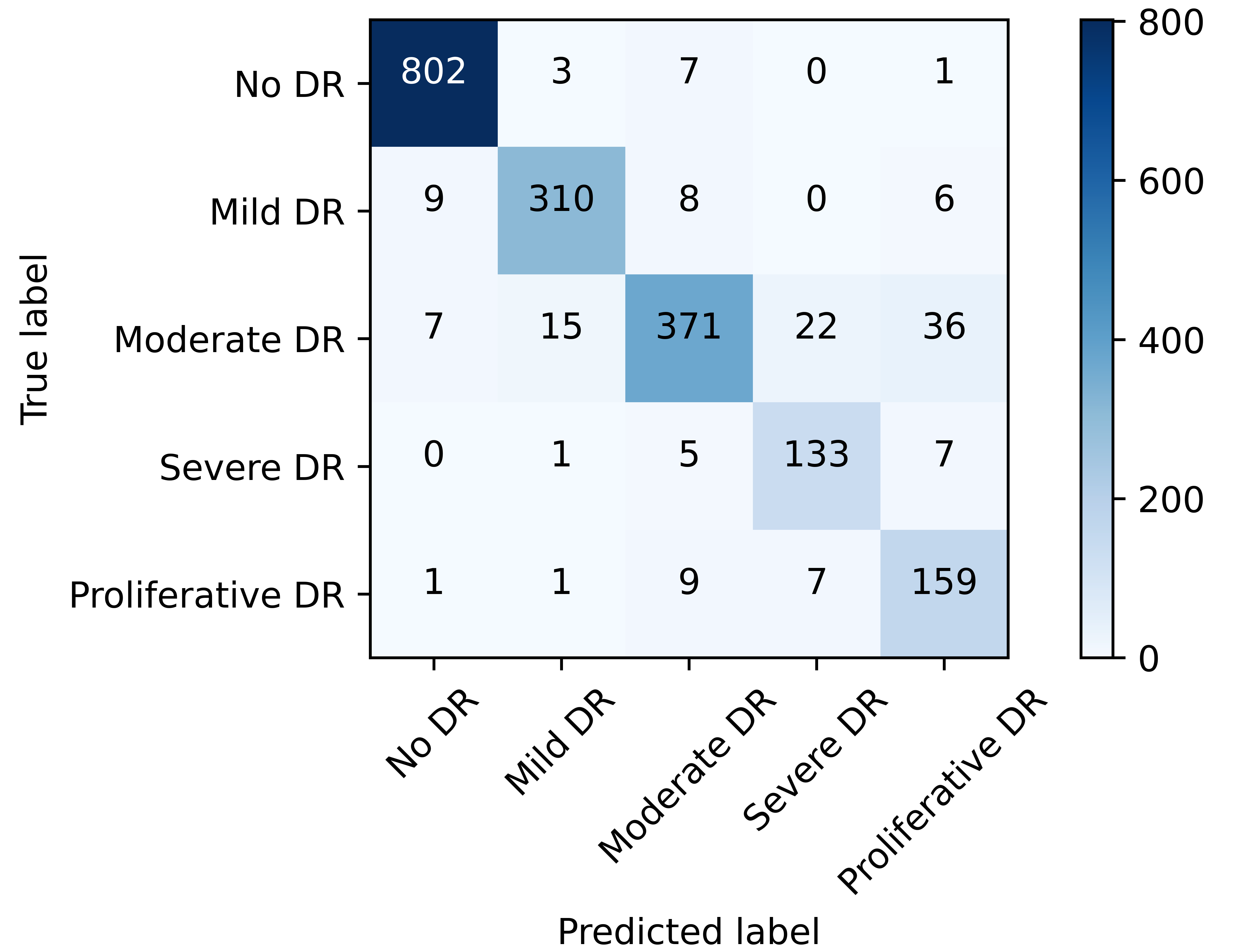}\label{fig:confm_cnn}}
    \subfigure[BCNN]{\includegraphics[width=0.45\textwidth]{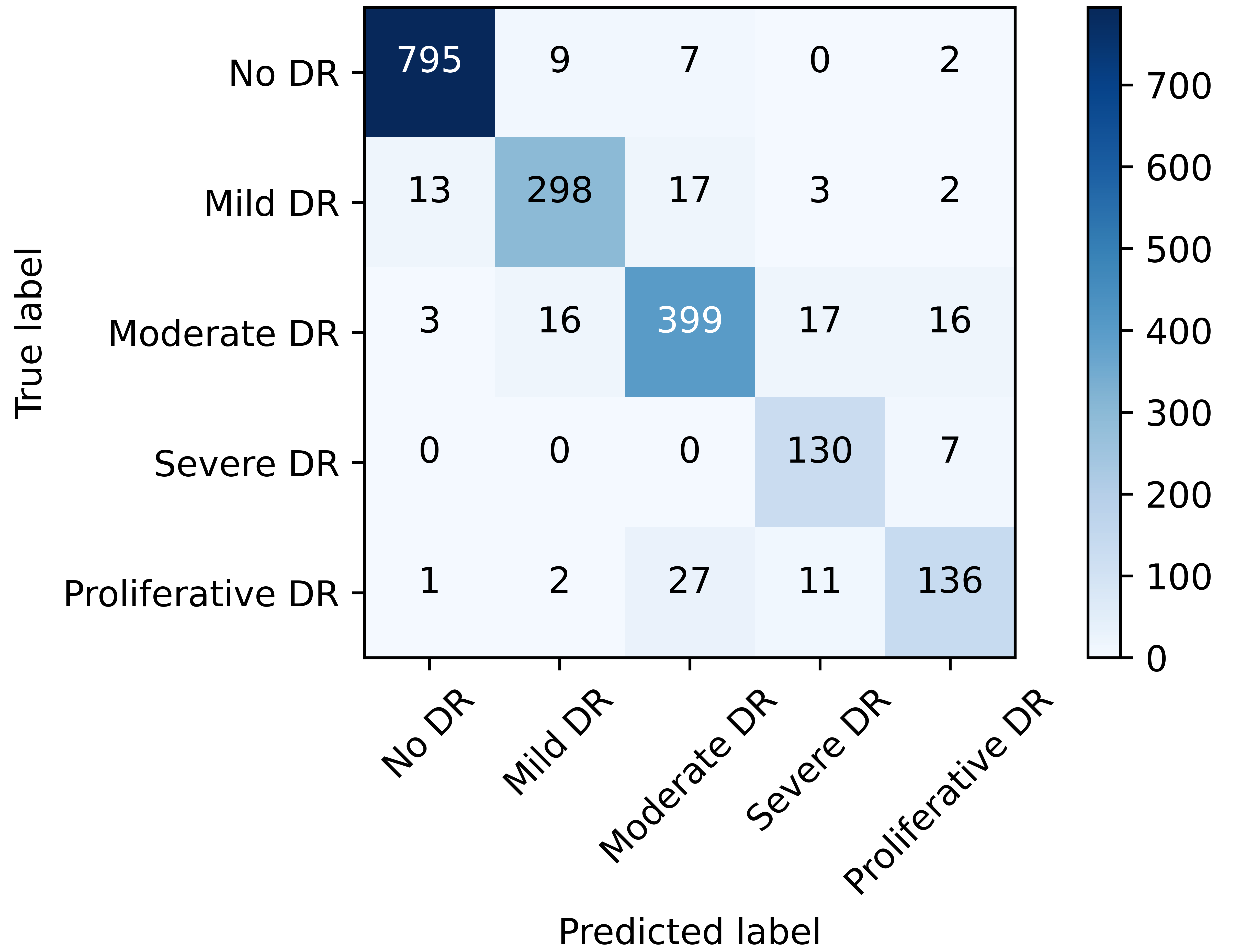}\label{fig:confm_mcdp}}
    \caption{Confusion matrices for multi-class classification.}
    \label{fig:confm_mc}
\end{figure*}

\begin{table}[!h]
\centering
\caption{Classification Report of CNN and BCNN in terms of accuracy, precision, recall and F1-score}
\label{tab:mc_cr}
\scalebox{0.76}{
\begin{tabular}{|c|c|c|c|c|c|c|c|}
\hline
\multicolumn{5}{|c|}{\textbf{Class 0/ No DR}} &
  \multicolumn{3}{c|}{\textbf{Class 1/ Mild DR}} \\ \hline
Model &
  \begin{tabular}[c]{@{}c@{}}Accuracy\\ (\%)\end{tabular} &
  \begin{tabular}[c]{@{}c@{}}Precision\\ (\%)\end{tabular} &
  \begin{tabular}[c]{@{}c@{}}Recall\\ (\%)\end{tabular} &
  \begin{tabular}[c]{@{}c@{}}F1-Score\\ (\%)\end{tabular} &
  \begin{tabular}[c]{@{}c@{}}Precision\\ (\%)\end{tabular} &
  \begin{tabular}[c]{@{}c@{}}Recall\\ (\%)\end{tabular} &
  \begin{tabular}[c]{@{}c@{}}F1-Score\\ (\%)\end{tabular} \\ \hline
BCNN &
  92 &
  92 &
  89 &
  91 &
  92 &
  89 &
  91 \\ \hline
CNN &
  92 &
  94 &
  93 &
  94 &
  94 &
  93 &
  94 \\ \hline
\multicolumn{5}{|c|}{\textbf{Class 2/ Moderate DR}} &
  \multicolumn{3}{c|}{\textbf{Class 3/ Severe DR}} \\ \hline
Model &
  \begin{tabular}[c]{@{}c@{}}Accuracy\\ (\%)\end{tabular} &
  \begin{tabular}[c]{@{}c@{}}Precision\\ (\%)\end{tabular} &
  \begin{tabular}[c]{@{}c@{}}Recall\\ (\%)\end{tabular} &
  \begin{tabular}[c]{@{}c@{}}F1-Score\\ (\%)\end{tabular} &
  \begin{tabular}[c]{@{}c@{}}Precision\\ (\%)\end{tabular} &
  \begin{tabular}[c]{@{}c@{}}Recall\\ (\%)\end{tabular} &
  \begin{tabular}[c]{@{}c@{}}F1-Score\\ (\%)\end{tabular} \\ \hline
BCNN &
  92 &
  87 &
  88 &
  88 &
  81 &
  89 &
  85 \\ \hline
CNN &
  92 &
  93 &
  82 &
  87 &
  82 &
  91 &
  86 \\ \hline
\multicolumn{8}{|c|}{\textbf{Class 4/ Proliferative DR}} \\ \hline
Model &
  Accuracy (\%) &
  \multicolumn{2}{c|}{Precision (\%)} &
  \multicolumn{2}{c|}{Recall (\%)} &
  \multicolumn{2}{l|}{F1-Score (\%)} \\ \hline
BCNN &
  92 &
  \multicolumn{2}{c|}{83} &
  \multicolumn{2}{c|}{77} &
  \multicolumn{2}{c|}{80} \\ \hline
CNN &
  92 &
  \multicolumn{2}{c|}{76} &
  \multicolumn{2}{c|}{90} &
  \multicolumn{2}{c|}{82} \\ \hline
\end{tabular}}
\end{table}

\subsubsection{Comparison of Active Learning Methods}
\begin{figure*}[!h]
    \centering
    \captionsetup{justification=centering}
    \subfigure[CNN]{\includegraphics[width=0.43\textwidth]{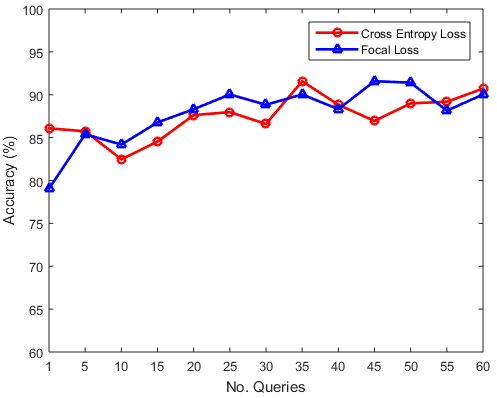}\label{fig:pbs_cnn_multi}}
    \subfigure[BCNN]{\includegraphics[width=0.43\textwidth]{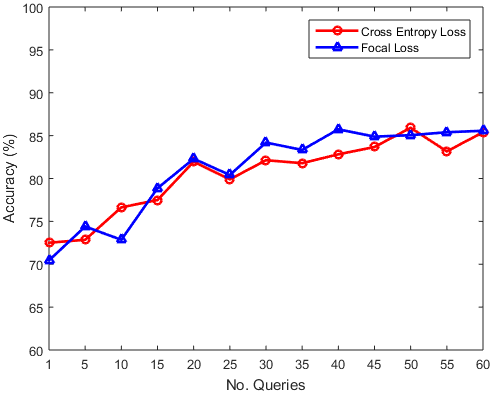}\label{fig:pbs_bcnn_multi}}
    \caption{Performance of \textbf{pool-based sampling} on the two loss functions (i.e., cross-entropy and focal loss.)}
    \label{fig:pbs_multi}
\end{figure*}

For multi-class classification, experiments have been separately performed on CNN and the BCNN. Both CNN and BCNN models initially achieved the performance of 78\% and 69\% when trained on a small number of dataset for $10$ epochs. The results of training a CNN model for pool-based sampling model $M_{PBS}$ are shown in Fig. \ref{fig:pbs_cnn_multi} and for BCNN are shown in Figure \ref{fig:pbs_bcnn_multi}. The final accuracy of the CNN and the BCNN model are 86\% and 85\%, respectively.


\begin{figure*}[!h]
    \centering
    \captionsetup{justification=centering}
    \subfigure[CNN]{\includegraphics[width=0.43\textwidth]{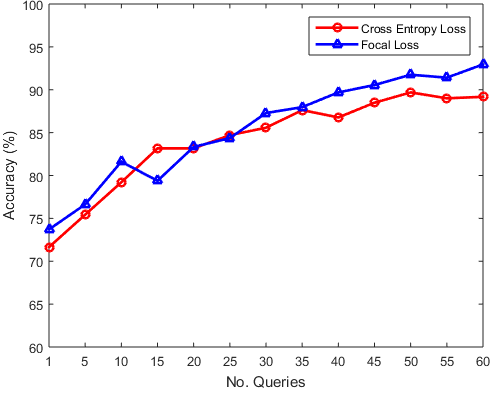}\label{fig:qbc_cnn_multi}}
    \subfigure[BCNN]{\includegraphics[width=0.43\textwidth]{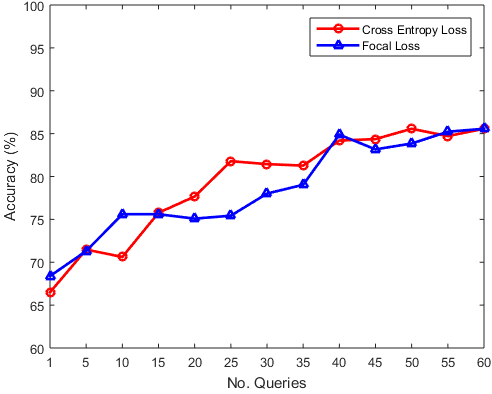}\label{fig:qbc_bcnn_multi}}
    \caption{Performance of \textbf{query-by-committee sampling} on the two loss functions (i.e., cross-entropy and focal loss.)}
    \label{fig:qbc_multi}
\end{figure*}

\begin{figure*}[!h]
    \centering
    \captionsetup{justification=centering}
    \subfigure[]{\includegraphics[width=0.43\textwidth]{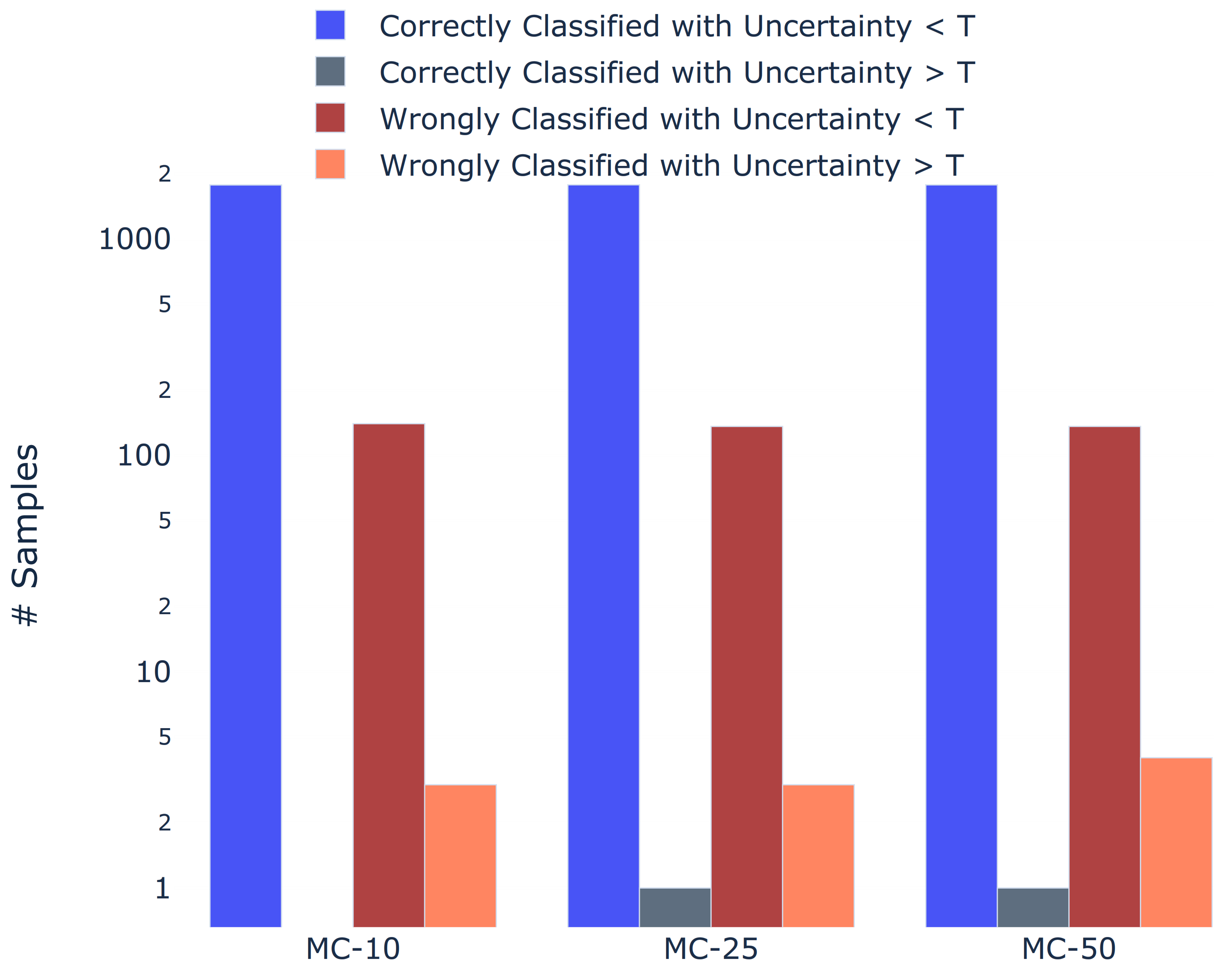}\label{fig:uq_bcnn_dp1}}
    \subfigure[]{\includegraphics[width=0.43\textwidth]{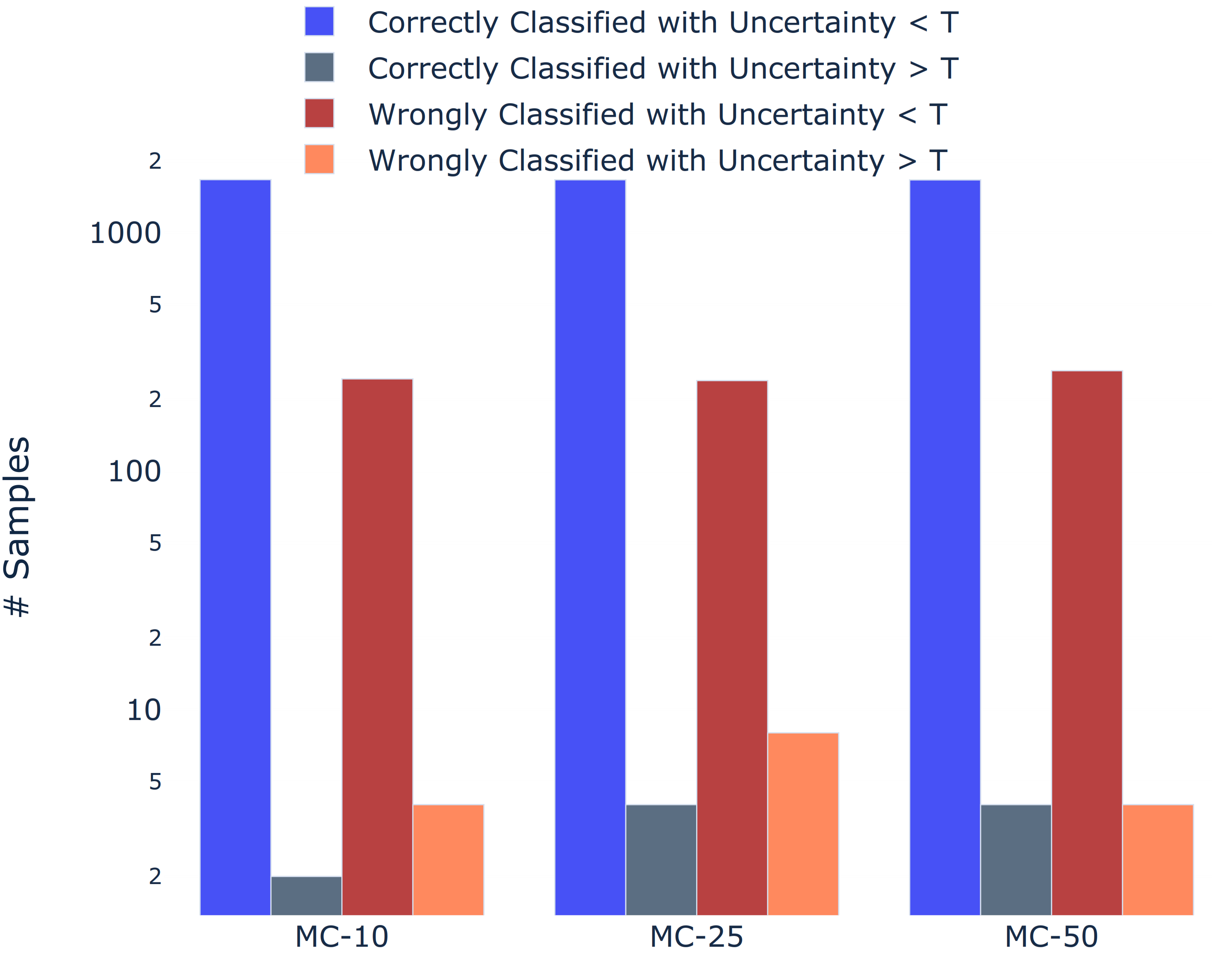}\label{fig:uq_bcnn_dp2}}
    \caption{ Uncertainty Quantification results on the BCNN models for the threshold value of $1.276$ and (a) dropout = $0.20$ and (b) dropout = $0.30$}
    \label{fig:uq_bcnn_dps}
\end{figure*}

\begin{table*}[!h]
\centering
\footnotesize
\caption{Comparison of the existing methods and with baseline for DR classification.}
\scalebox{0.8}{\begin{tabular}{|c|c|l|c|c|}
\hline
Author &
  Year &
  \multicolumn{1}{c|}{Methods} &
  Dataset(s) &
  Results \\ \hline
Ours &
  2020 &
  \begin{tabular}[c]{@{}l@{}}- Monte-Carlo Dropout\\ - Hybrid Model\\ - Simple active learning query functions\\ - Uncertainty quantification\\ - Binary and multi-class classification\\ - Automated method  of labeling  unlabeled data\end{tabular} &
  APTOS 2019 &
  \begin{tabular}[c]{@{}c@{}}AUC = 0.99 \\ (multi-class classification)\\ Accuracy = 92\% \\ (multi-class classification )\\ Accuracy = 85\% \\ (BCNN in  Active  Learning)\end{tabular} \\ \hline
Chetoui at el. \cite{chetoui2020explainable} &
  2020 &
  \begin{tabular}[c]{@{}l@{}}- EfficientNet-B7 model\\ - Binary classification (referred DR/visual-threatening DR)\\ - Gradient-weighted Class Activation Mapping (Grad-CAM) to detect signs of DR\end{tabular} &
  \begin{tabular}[c]{@{}c@{}}APTOS 2019\\ EyePAC 2015\\ (Kaggle)\end{tabular} &
  \begin{tabular}[c]{@{}c@{}}AUC = 0.996 \\ (for referred DR)\\ AUC = 0.998 \\ (for visual threatening DR) \\ Accuracy = not reported\end{tabular} \\ \hline
\multicolumn{1}{|l|}{Khalifa at el. \cite{khalifa2019deep}} &
  \multicolumn{1}{l|}{2019} &
  \begin{tabular}[c]{@{}l@{}}- Data augmentation (horizontal flip/vertical flip) \\- AlexNet, ResNet, VGG16/19, SqueezeNet and GoogleNet\\ - Transfer learning for DR classification\end{tabular} &
  APTOS 2019 &
  \begin{tabular}[c]{@{}c@{}}\\Accuracy = 97.9\%\\ \\ AUC = not reported \\ \end{tabular} \\ \hline
Filos et al. \cite{filos2019systematic}&
  2019 &
  \begin{tabular}[c]{@{}l@{}}- Mean-field variational inference\\ - Monte-Carlo Dropout\\ - Deep Ensembles\\ - Uncertainty quantification\\ - Binary class classification only\end{tabular} &
  \begin{tabular}[c]{@{}c@{}}EyePAC 2015\\ (Kaggle)\end{tabular} &
  \begin{tabular}[c]{@{}c@{}}Accuracy = 84\% \\ (No referral)\\ \\ Accuracy = 91.3\% \\ (50\% data referred)\end{tabular} \\ \hline
Lam et al. \cite{lam2018automated} &
  2018 &
  \begin{tabular}[c]{@{}l@{}}- Contrast limited adaptive histogram equalization\\ - GoogleNet / AlexNet\\ - Transfer learning with  ImageNet  weights\end{tabular} &
  \begin{tabular}[c]{@{}c@{}}EyePAC 2015\\ (Kaggle)\end{tabular} &
  \begin{tabular}[c]{@{}c@{}}Accuracy = 75\%\\ AUC = not reported\end{tabular} \\ \hline
Gal et al. \cite{gal2017deep}&
  2017 &
  \begin{tabular}[c]{@{}l@{}}- Monte-Carlo Dropout\\ - Training of Bayesian CNN model in active learning settings\\ - Customized query functions\end{tabular} &
  \begin{tabular}[c]{@{}c@{}}MNIST\\ IPIC 2016\end{tabular} &
  \begin{tabular}[c]{@{}c@{}}AUC = 0.75\\ Accuracy = not reported\end{tabular} \\ \hline
\end{tabular}}
\label{tab:comp}
\end{table*}

Similarly, for query-by-committee model $M_{QBC}$, the result of CNN are shown in Figure \ref{fig:qbc_cnn_multi} and for BCNN are shown in Figure \ref{fig:qbc_bcnn_multi}. $M_{QBC}$ for the CNN model has a higher value of accuracy in both cross-entropy and in the focal loss. 

The results are showing that the focal loss is performing better for active learning models, i.e., $M_{PBS}$ and $M_{QBC}$. In comparison with overall performance, the focal loss in BCNN for $M_{QBC}$ is performing best as its learning behavior is comparatively smooth. As the new data is being added, all three committee models are learning more from the hard examples. Also, the focal loss is enforcing the model to not get over-fitted on the already known data and focusing on the newly augmented examples by taking them as hard examples to train. 

\subsubsection{UQ for Multi-class Classification}
For the UQ, we use two different dropout rates of $0.20$ and $0.30$. As in the binary class classification, we use different MC samples and reported the correctly and wrongly classified samples. We set the threshold value as $T$ of $1.276$. The sample whose entropy is less than $T$ and whose prediction is the same as the ground truth label is counted as correctly classified. The UQ results are reported only on the BCNN model. The results for the two dropout rates, i.e., $0.20$ and $0.30$ are shown in Figure \ref{fig:uq_bcnn_dp1} and in Figure \ref{fig:uq_bcnn_dp2}, respectively. The \textit{x-axis} is showing the MC iterations and the \textit{y-axis} is the \textit{log-scale} of total number of samples. It can be seen from both figures that increasing the dropout rate in BCNN is reducing the performance of correctly classified samples with uncertainty less than the Threshold value.

\subsection{Comparison with Existing Methods}
\label{sub:res_comp_em}
We compare our approach with two approaches independently. Firstly, we compare the performance of MC dropout with our baseline paper \cite{filos2019systematic} and extended their approach to the multi-class classification by training our models with less number of training data samples. Secondly, we compared our proposed method of training a hybrid model (Bayesian) in active learning settings with \cite{gal2017deep}. As we used classical AL query strategies like \textit{uncertainty sampling} and \textit{vote-entropy sampling} which are quite intuitive and straightforward querying methods of obtaining the most informative samples. The overall comparison with the existing methods can be seen in Table \ref{tab:comp}.
\subsection{Discussions}
\label{ref:sub:disc}
We now discuss our analysis and identify some of the interesting insights and a few points which can be considered for future work.

In conventional AL approaches, the required label of queried samples is obtained manually either by asking from an expert field annotator or its ground truth is already available for training. We replaced this approach by automating the process of acquiring the label from a well trained BCNN model. We only forward those labels on which our annotator model (which we call feature descriptor) is quite confident (the case where uncertainty is less than the threshold). Still there is a risk that a wrongly classified example with more confidence can mislead the AL models to wrong learning. For now, we have cross verified our approach by adding only those samples in AL settings on which our ground truth label is same as the predicted label and the uncertainty is less than the specified threshold value. 

The threshold used for uncertainty and selection of dropout parameters is among the important parameters in training and evaluating the MC dropout approach. There are a lot of hyper-parameters involved in training and evaluating the complex and large-sized CNN models. We reported our best results, but still we believe that further optimization of the hyperparameters can be performed to achieve more stable results. Furthermore, in medical imaging, we need to quantify uncertainty and we need to incorporate further statistical approaches that need to be investigated. In addition, we would like to note that \textit{MC dropout} is not the only way of approximating the true posterior distribution. There are other methods like variational inference specifically mean-field variational inference which approximates the posterior distribution by minimizing the KL-divergence between the two distribution which can be investigated for the task of DR classification (especially in multi-class classification).

The complexity of neural network models is always a challenging issue and problem-specific neural networks need to be designed. Most of the DL models for medical image diagnosing use the phenomenon of \textit{transfer learning} and use the pre-trained weights of \textit{ImageNet} to fine-tune the models. Recent studies \cite{raghu2019transfusion} have revealed that techniques like \textit{transfer learning} offers limited performance for medical imaging tasks using the weights of \textit{ImageNet}. By keeping this in mind, we designed all of our models as independent of transfer learning and they are being trained from scratch using their own randomly initialized weights.

Lastly, we also encourage the interested readers to think about multi-label classification (i.e., assigning more than one label at a time to a single sample) using Bayesian CNNs and investigate the uncertainty quantification for this task. This approach can help the medical experts to see the possible transition from one stage to another stage and can help them to wisely suggest the related therapies.

\section{Conclusions}
\label{sec:conclusion}
In this paper, we have proposed a hybrid model for the problem of diabetic retinopathy (DR) classification that jointly handles uncertainty problem and is also able to learn from unlabelled data. In particular, the proposed framework has two main components, i.e., the Bayesian convolutional neural network (BCNN) model having Monte-Carlo drop-out, which is used as a feature descriptor and an active learning (AL) component. BCNN reduced the uncertainty of the prediction, while the AL module enables learning from unlabelled data. We have performed an extensive evaluation of the proposed framework under different settings and also, we have compared the performance of BCNN with that of deterministic CNN. We have evaluated our approach for both binary class classification and multi-class classification and have achieved competitive results as compared to the state-of-the-art. Our BCNN model for binary class classification has achieved an accuracy of $92\%$ (less confident) and $81\%$ (more confident), while our multi-class BCNN model has achieved an accuracy of $92\%$ (more confident). Moreover, our AL results for the BCNN model for binary class classification are improving state-of-the-art results with an accuracy of $91\%$ and for the case of multi-class classification, AL models for both CNN and BCNN needs to be optimized further in future work.

\bibliographystyle{IEEEtran}

\end{document}